\documentclass[conference]{IEEEtran}
\IEEEoverridecommandlockouts
\usepackage{cite}
\usepackage{url}
\usepackage{CJK}
\usepackage{amsfonts}
\usepackage{graphicx}
\usepackage{textcomp}
\usepackage{xcolor}
\usepackage{algorithm}
\usepackage{algorithmicx}
\usepackage{algpseudocode}
\usepackage{amsmath}
\usepackage{amssymb}
\usepackage{graphics}
\usepackage{epsfig}
\usepackage{indentfirst}
\usepackage{multirow}

\floatname{algorithm}{Algorithm}

\def\BibTeX{{\rm B\kern-.05em{\sc i\kern-.025em b}\kern-.08em
    T\kern-.1667em\lower.7ex\hbox{E}\kern-.125emX}}
\begin{document}

\title{Self-attention-based BiGRU and capsule network for named entity recognition*\\
{\footnotesize \textsuperscript{*}Note: Sub-titles are not captured in Xplore and
should not be used}
\thanks{Identify applicable funding agency here. If none, delete this.}
}

\author{\IEEEauthorblockN{Jianfeng Deng}
\IEEEauthorblockA{\textit{School of Automation} \\
\textit{Guangdong University of Technology}\\
Guangzhou, China \\
dengjianfeng0839@gmail.com}
\and
\IEEEauthorblockN{Lianglun Cheng}
\IEEEauthorblockA{\textit{School of Computer} \\
\textit{Guangdong University of Technology}\\
Guangzhou, China \\
llcheng@gdut.edu.cn}
\and
\IEEEauthorblockN{Zhuowei Wang}
\IEEEauthorblockA{\textit{School of Computer} \\
\textit{Guangdong University of Technology}\\
Guangzhou, China \\
wangzhuowei0710@163.com}
}

\maketitle

\begin{abstract}
Named entity recognition(NER) is one of the tasks of natural language processing(NLP). In view of the problem that the traditional character representation ability is weak and the neural network method is unable to capture the important sequence information. An self-attention-based bidirectional gated recurrent unit(BiGRU) and capsule network(CapsNet) for NER is proposed. This model generates character vectors through bidirectional encoder representation of transformers(BERT) pre-trained model. BiGRU is used to capture sequence context features, and self-attention mechanism is proposed to give different focus on the information captured by hidden layer of BiGRU. Finally, we propose to use CapsNet for entity recognition. We evaluated the recognition performance of the model on two datasets. Experimental results show that the model has better performance without relying on external dictionary information.
\end{abstract}


\section{Introduction}
NER is an important task in natural language processing. The purpose is to identify specific entity information in the text\cite{nadeau2007survey}. Generic domain entities include names of people, places, times, and organizations\cite{ZhangHainan2017}. Specific areas such as medical fields, entities include disease names, patient symptoms, and treatment options\cite{li2018survey}. With the rapid development of electronic information, a large number of unstructured texts accumulate over time. Massive text data contains a lot of semantic knowledge. In order to fully mine the valuable information in the unstructured text and lay a good data foundation for the next knowledge graph construction, intelligent question, answer system and other basic applications\cite{liuqiao2016}, NER is an essential step.

In the past, long-term entity recognition work mainly focused on texts written in English. With the large increase in the number of Chinese texts, it is urgent to extract useful information from Chinese texts. However, due to the following reasons, the entity recognition task still faces challenges: 

(1)Compared with English, Chinese lacks explicit lexical boundaries and inherent definite articles, and proprietary words have no prompting information such as spelling changes; 

(2)The supervised Chinese training data are limited; 

(3)There is no unified language format for domain texts and entities has problems such as combination and abbreviation. 

Currently, deep learning is mainly used for NER research. These methods use neural networks to capture word feature information from a sequence of word vectors for entity recognition. Compared with traditional machine learning and rule matching methods, deep learning models have excellent performance, but still have challenges. There are two kinds of deep learning models in NER: Convolutional Neural Network (CNN)\cite{collobert2011natural} and Recurrent Neural Network (RNN)\cite{zhou2017joint}. CNN can capture local salient features of text, but lacks the ability to capture context features. RNN processes sequence data sequentially, so it can capture the context features of text sequences, but non-entity words occupy most of the sentence, and a lot of redundant information participates in the entity recognition process. Redundant information hinders the acquisition of important entity feature information, leading to decrease in the accuracy of entity recognition. When CNN or RNN is used for entity recognition, using feature weighted sum to represent entity label probability. This process loses most of the data information, which leads to incorrect judgment when facing the same word at different positions in the sequence.

In order to solve the above problems, this paper proposes a new NER model called B-SABCN, which aims to enhance the model's ability to capture important information. Firstly, the BERT pre-trained model is used to train the character embedded representation, which can improve the expression ability of character vector information. Secondly, the BiGRU network is used to capture the context information. For the problem of information redundancy, a self-attention mechanism is proposed to give different focus on features captured by the hidden layer of BiGRU. Finally, it is proposed to use CapsNet to classify entity categories in order to obtain richer information to generate entity label sequence. This paper evaluates the performance of the B-SABCN model on the general domain Chinese NER dataset and the domain specific Chinese NER dataset. Compared with advanced methods in recent years. Experimental results show that the model has good performance without providing external information.

The main contents of this paper are as follows. The section 2 introduces the literature review of NER. In section 3, the model for entity recognition task is introduced in detail. Section 4 is the experimental results and results analysis of the model. Finally, section 5 gives some conclusions and possible approaches for future research. 

\section{Related work}
NER can be viewed as a sequence labeling problem. The main methods can be divided into three categories: rule-based\cite{Zhangyanli2001}, machine learning-based\cite{chieu2002named} and deep learning-based. Among them, the deep learning-based method can automatically capture the text features from the input feature sequence and achieve end-to-end entity recognition, which is a current research hotspot.
\subsection{Input feature representation}
In the Chinese NER task, Chinese has two forms of characters and words. Characters and words have corresponding semantics. Wang et al\cite{wang2017named}. proposed to generate feature vectors at the character level and build a Chinese NER model. The results show that the accuracy of the character-based entity recognition method is higher than the word-based entity recognition method. In addition to character feature vector representation, additional features can be added as a way to enrich character feature information. He et al\cite{he2016f}. uses word feature information as a supplement to character features to enhance recognition. Dong et al\cite{dong2016character}. uses character radical features and character features as input features, which can solve the problem of weak traditional character feature vector representation ability to a certain extent. These experiments have improved the accuracy of entity recognition.

However, the character or word vector of the above method is generated by the word2vec\cite{mikolov2013distributed} method. Word2vec trains character vectors based on neural networks, and word vectors are fixed. In Chinese semantics, characters in different words have different meanings, which contradicts a fixed word vector. Although word features and character radical features can enrich character features, this features needs to be provided by external resources and is not suitable for all Chinese entity recognition data sets. Aiming at the inability of traditional methods to express character ambiguity, researchers have proposed a new character vector pre-training model that can improve character feature representation without providing external information. Matthew et al\cite{peters2018deep}. proposed embeddings from languages models(ELMO) to learn lexical polysemy in different contexts. Devlin et al\cite{devlin2018bert}. proposed BERT model and used a bidirectional transformer as an encoder, which achieved good results in NLP tasks. Therefore, this paper uses a BERT pre-trained language model to train Chinese character vectors.
\subsection{Text features capture}
In recent years, CNN and RNN are commonly models used for capturing text features. Santos et al\cite{santos2015boosting}. use the joint representation of letters and words as input to CNN model for entity recognition. Since CNN is constrained by convolution kernels of a fixed size. Strubell et al\cite{strubell2017fast}. use an iterative inflated convolutional neural network(ID-CNN) to capture a wide range of information by adding convolution kernels. Compared with CNN, RNN is usually used for sequential modeling, and it is more common than CNN application in NER tasks. Hammerton et al\cite{hammerton2003named}. first proposed NER model based on long short-term memory network(LSTM). LSTM can overcome gradient explosion and vanlishing problems in RNN. Later researchers improved LSTM and proposed that bidirectional long short-term memory network(BiLSTM)\cite{ma2016end}. BiLSTM consists of forward LSTM unit and backward LSTM unit to enhance the ability of neural networks to capture context information. Huang et al\cite{huang2015bidirectional}. proposed a BiLSTM-based NER, and the results showed that the recognition effect based on BiLSTM was better than that based on LSTM. Katiyar et al\cite{katiyar2018nested}. proposed multi-layer BiLSTM to address the problem of nested NER. Therefore, BiLSTM becomes the basic structure of entity recognition.

The RNN is divided into two variants, LSTM and GRU\cite{tang2016question}. LSTM unit processes text sequences through input gate, forget gate, and output gate. GRU unit controls information transmission through input gate and update gate. Compared with LSTM, GRU has fewer parameters and has a simple structure. It can shorten the training time when the performance is comparable. This paper uses BiGRU to capture text context information.
\subsection{Attention mechanism}
After the neural network model captures text features, the contribution of each character to the NER remains the same. In fact, there are a large number of non-entity characters in the sentence. Some characters can play an important role in the NER task. These important information can be assigned a larger weight. Attention mechanisms have become a way to learn important information about sequences. Pandey et al\cite{pandey2017improving}. proposed BiLSTM model with attention mechanism to identify medical entities from electronic health records. Luo et al\cite{luo2018attention}. used attention mechanisms for different sentences to solve the problem of marking inconsistencies. It achieved state-of-the-art results at that time. 

However, most attention mechanisms calculate the relationship coefficient between characters and entity labels as character feature weights. It is hard to capture useful information for entity recognition between characters. Although BiLSTM can capture information between characters, it is difficult for BiLSTM to capture information between characters at long distances in text sequences. Vaswani et al\cite{vaswani2017attention}. proposed a self-attention mechanism that can learn the dependency relationship between arbitrary characters and overcome the problem of BiLSTM capturing information between long distances characters. The information between characters can further understand the sentence structure and improve the performance of entity recognition. In this paper, the self-attention mechanism is used to further obtain the relevant features of text itself from multiple spaces on the basis of the features captured by BiGRU.
\subsection{Entity category recognition}
When using CNN or BiLSTM for entity recognition, scalar neurons are used to represent the probability of entity labels, and the ability to express entity information is weak. In contrast, Hinton et al\cite{sabour2017dynamic}. firstly proposed a CapsNet in 2017, which is different from the scalar neuron output of a neural network. The CapsNet outputs capsule vector that represents each category. Zhao et al\cite{zhao2018investigating}. proposed CapsNet for text classification. The results show that the classification performance is improved. In this paper, CapsNet is introduced to predict entity labels. Capsules represent entity labels, and capsule vector modulus length represents entity label probabilities. Using vectors instead of scalar neurons can represent richer entity information.

In the NER task, there is a dependency on the entity recognition between characters. The entity recognition of one character affects the recognition of the next character. Therefore, a constraint method needs to be added to ensure the recognition result is reasonable. Common constraint methods include Hidden Markov Model(HMM)\cite{YuHongkui2006}, Conditional Random Field(CRF)\cite{SunLiping2016}, and so on. Among them, CRF is an improved HMM model that can learn context information sequentially, so it has become a mainstream constraint method.

Currently, the BiLSTM-CRF model is a commonly used model for Chinese NER. Its performance depends on the input feature representation and the ability of the neural network to capture information. The existing NER method is still challenging, so this paper proposes a B-SABCN model for the Chinese NER task. It can effectively represent character feature representations and enhance the ability of neural networks to capture important information. The B-SABCN is evaluated on two data sets, and the results show that the B-SABCN model achieves the best recognition effect.
\section{Model}
This section describes the structure of the B-SABCN model, as shown in figure 1. The B-SABCN model consists of three parts: input feature layer, BiGRU and self-attention mechanism layer, and CapsNet entity recognition layer. The input feature layer is mainly represented by BERT pre-trained model training character vector. The BiGRU network mainly captures the text context features, and the self-attention mechanism further acquires the deep features of the text. CapsNet predicts the entity label corresponding to the character, and then adds a Markov transfer function to constrain the prediction result to achieve entity recognition. The following sections detail the component structure of the model.
\begin{figure}[htbp]
\centering\includegraphics[width=0.58\textwidth]{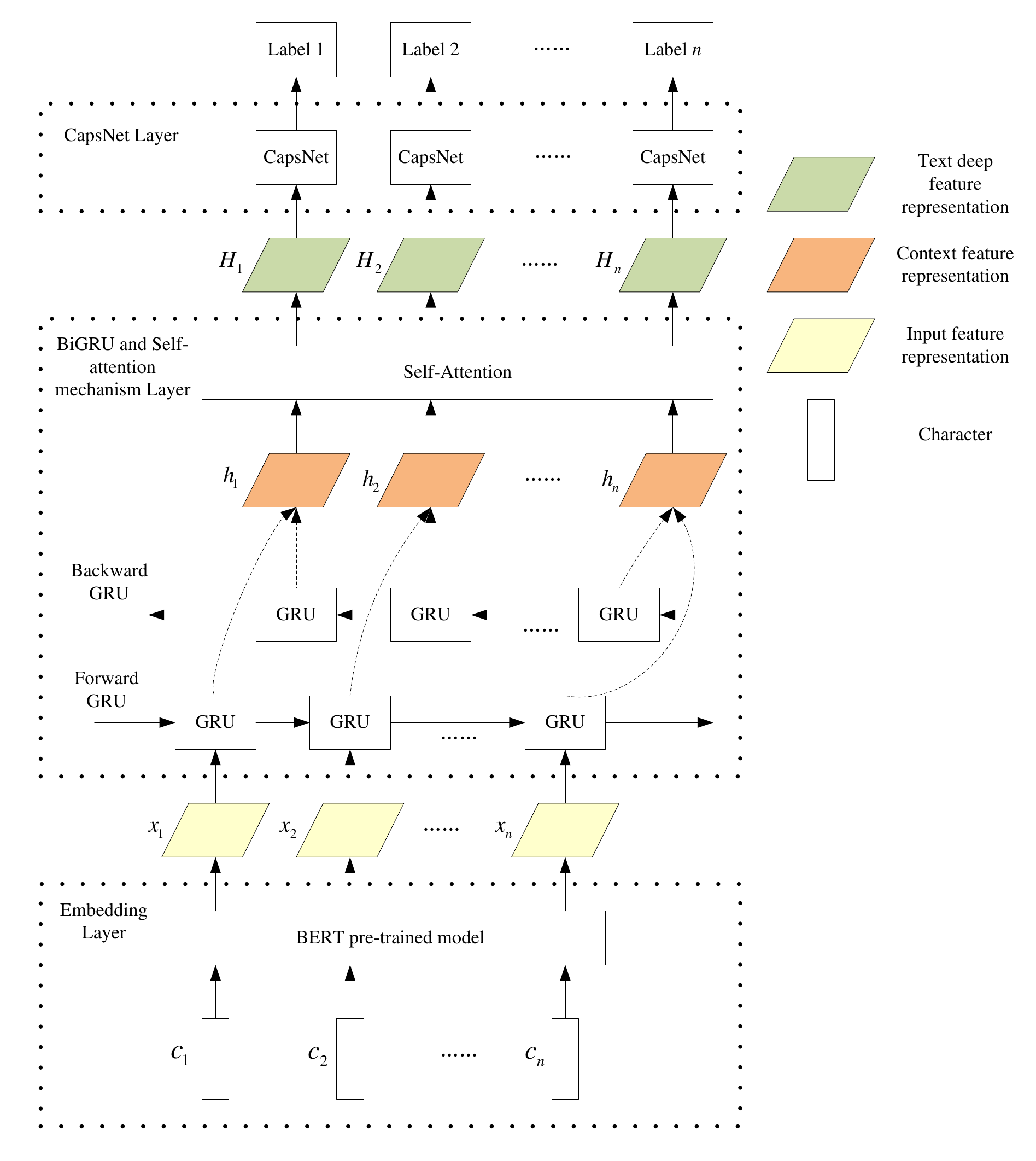}
\caption{The overall structure of our proposed B-SABCN model}
\label{fig1}
\end{figure}
\subsection{Embedding Layer}\label{AA}
BERT model uses bidirectional Transformers as encoder, fusing the context information of the left and right sides of the current character. When training character vectors, the encoder no longer encodes sentences from left to right or from right to left to predict characters, but randomly hides or replaces some characters according to a certain proportion and predicts the original characters according to the context. In addition, the BERT model also adds sentence level training tasks, mainly learning the context relationship between sentences. The specific practice is to replace some sentences randomly, and the encoder uses the previous sentence to predict whether the next sentence is the original sentence. These two tasks are jointly trained to capture vector representations at the character level and sentence level.

In this paper, given a sequence of sentences. BERT uses the fine-tuning parameter mechanism. The input sequence is set to $c = ([CLS], c_{1}, c_{2},... , c_{n}, [SEP])$, where $[CLS]$ represents the beginning of a sample sentence sequence, and $[SEP]$ represents the spacing symbol between sentences. They are used for sentence-level training tasks. The representation of each character vector consists of three parts: character embedding vector, sentence embedding vector and position embedding vector. The character embedding vector is defined as ${e^c} = (e_{[CLS]}^c, e_{{c_1}}^c, e_{{c_2}}^c,... , e_{{c_n}}^c, e_{[SEP]}^c)$, the sentence embedding vector is defined as ${e^s} = (e_{A}^s, e_{A}^s, e_{A}^s,... , e_{A}^s, e_{A}^s)$, and the character position embedding vector is defined as ${e^p} = (e_{0}^p, e_{1}^p, e_{2}^p,... , e_{n}^p, e_{n+1}^p)$. Among them, the character embedding vector is determined by the vocabulary provided by BERT. Because the training sample is a single sentence, the sentence embedding vector is set to 0. Add the three embedding vectors together to get character feature as input of BERT model, as shown in figure 2. Character vector input through training, the final character vector representation is shown in formula 1, as input to BiGRU and self-attention mechanism layer.
\begin{equation}
x =
\left[
\begin{array}{ccccc}
x_{1},x_{2},...,x_{n}\\
\end{array}
\right]
\end{equation}
\begin{figure}[htbp]
\centering\includegraphics[width=0.45\textwidth]{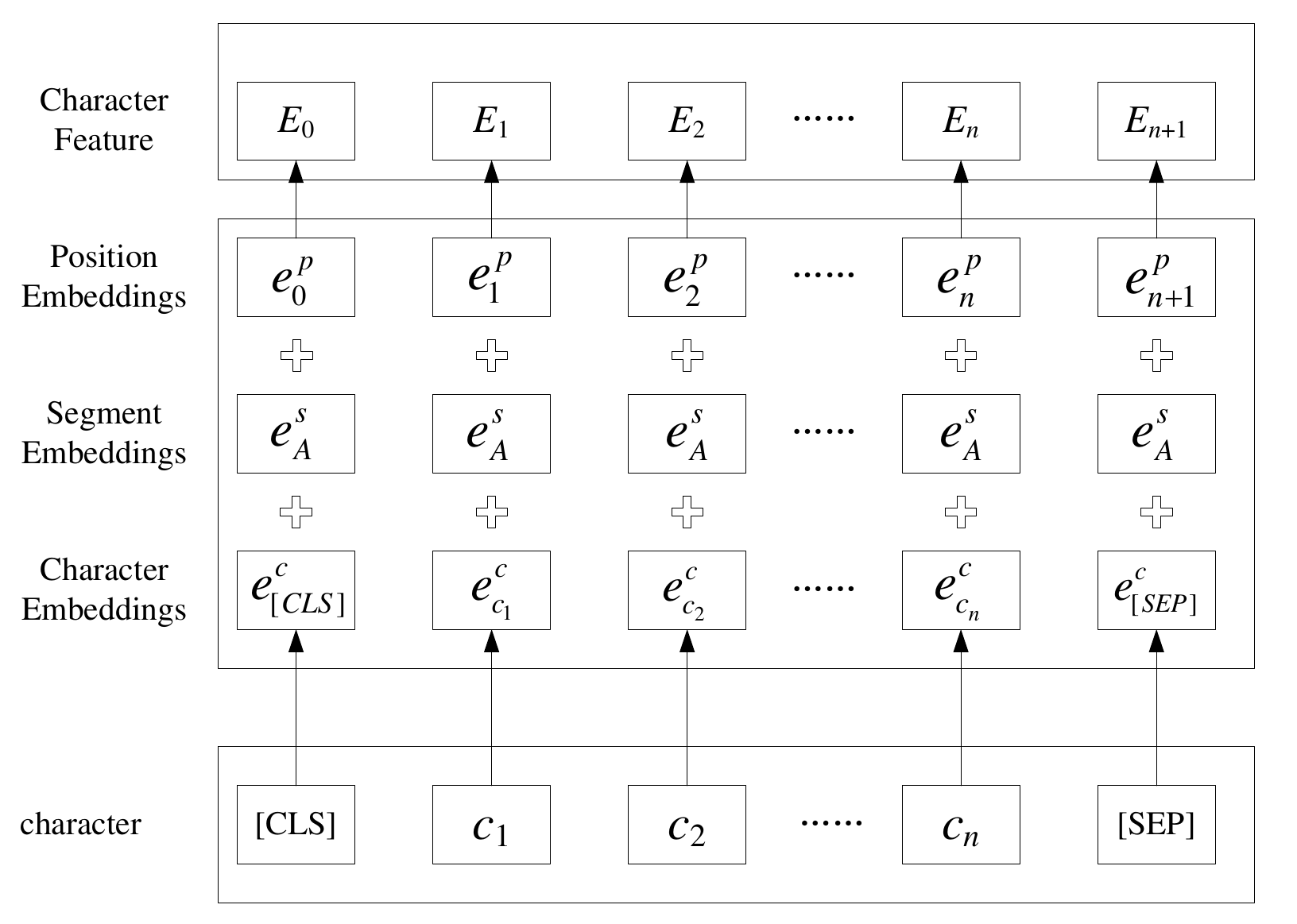}
\caption{Character vector representation input based on BERT model}
\label{fig2}
\end{figure}

\subsection{BiGRU and Self-attention mechanism Layer}
Intuitively, entity recognition is the processing of sentence sequence information. BiGRU is used in sequence modeling to extract context information from character vector sequences. BiGRU consist of forward GRU unit and backward GRU unit. GRU consist of update gate and reset gate. The gate structure can choose to save context information to solve the problem of RNN gradient disappearance or explosion. The GRU cell is shown in figure 3.
\begin{figure}[htbp]
\centering\includegraphics[width=0.4\textwidth]{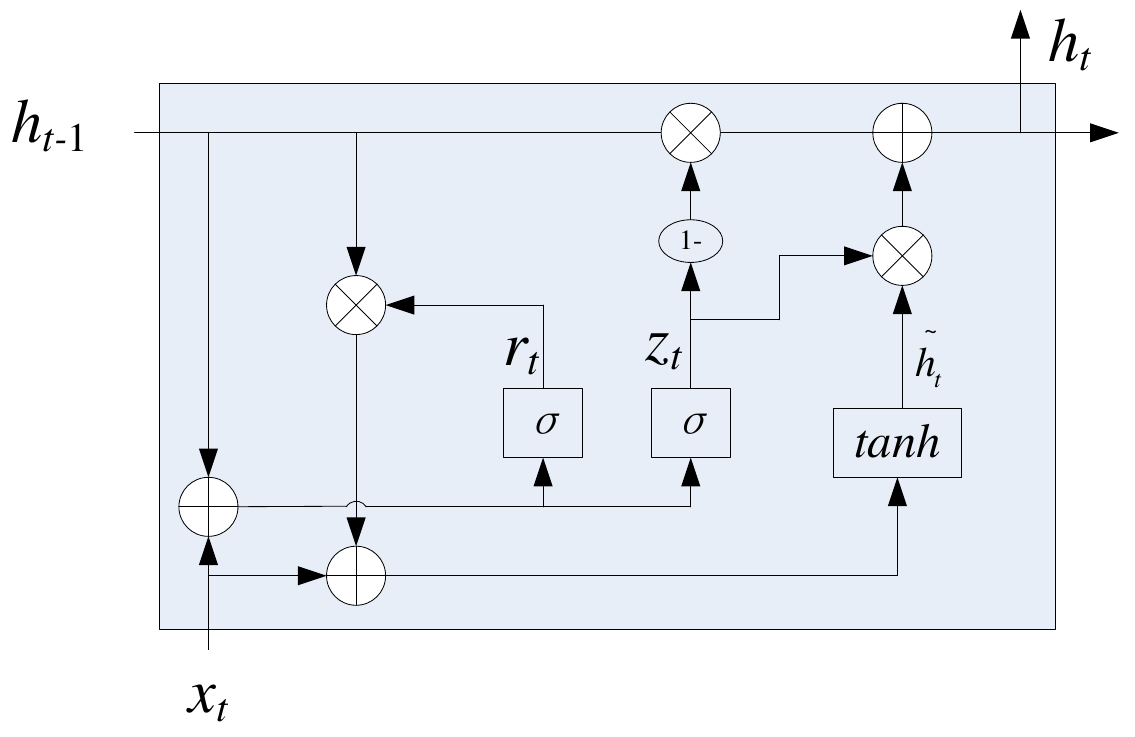}
\caption{GRU unit structure}
\label{fig3}
\end{figure}

For time $t$, the GRU cell state calculation formula is as follows:
\begin{eqnarray}
{r_t} = \sigma ({w_r} \cdot [{h_{t - 1}},{x_t}] + {b_r})
\end{eqnarray}
\begin{eqnarray}
{z_t} = \sigma ({w_z} \cdot [{h_{t - 1}},{x_t}] + {b_z})
\end{eqnarray}
\begin{eqnarray}
\tilde{h}_t = \tanh ({w_h} \cdot [{r_t}*{h_{t - 1}},{x_t}] + {b_h})
\end{eqnarray}
\begin{eqnarray}
{h_t} = (1 - {z_t})*{h_{t - 1}} + {z_t}*\tilde{h}_t
\end{eqnarray}

Where $\sigma$ is the sigmoid function and $\cdot$ is the dot product. ${w_r}$, ${w_z}$, ${w_h}$ are weight matrices and ${b_r}$, ${b_z}$, ${b_h}$ are bias parameters. ${x_t}$ is the input vector of time $t$, ${h_t}$ is the hidden state, and is also the output vector, containing all the valid information of the previous $t$ time. ${z_t}$ is an update gate, which is used to control the influence of the previous unit's hidden layer output on the current unit state. The larger value of the update gate, the greater influence of the previous unit's hidden layer output on the current unit state. ${r_t}$ is a reset gate, which is used to control the importance of ${h_{t - 1}}$ to $\tilde{h}_t$. The smaller value of the reset gate, the greater degree to which the hidden layer information of the previous unit is ignored. $\tilde{h}_t$ represents information that needs to be updated in current unit. Both gates capture sequence length dependence. In the case that the performance is equivalent to LSTM, the structure of GRU is simpler than LSTM and the training speed is faster.

The BiGRU network adopted in this paper is composed of forward GRU unit and backward GRU unit. The hidden layer of forward GRU unit is expressed as $\mathop {{h_t}}\limits^ \to$ and the hidden layer of backward GRU unit is expressed as $\mathop {{h_t}}\limits^ \leftarrow$. Through formula (2)-(5), the hidden layer outputs of unidirectional GRU at time $t$ are obtained, as shown in formula (6)-(7). The hidden layer output of BiGRU at time $t$ is spliced through the hidden layer output of forward GRU unit and backward GRU unit, as shown in formula 8.
\begin{eqnarray}
\mathop {{h_t}}\limits^ \to   = GRU({x_t},\mathop {{h_{t - 1}}}\limits^ \to)
\end{eqnarray}
\begin{eqnarray}
\mathop {{h_t}}\limits^ \leftarrow   = GRU({x_t},\mathop {{h_{t - 1}}}\limits^ \leftarrow)
\end{eqnarray}
\begin{eqnarray}
{h_t} = [\mathop {{h_t}}\limits^ \to  ,\mathop {{h_t}}\limits^ \leftarrow]
\end{eqnarray}

The purpose of BiGRU is to capture the context features of sentence sequence. However, in fact, the semantic information of each character in the sentence has different contributions to NER task. There is a large amount of useless information in the text, which result in information redundancy. BiGRU model is difficult to capture important information from sentence sequences. Therefore, after the BiGRU network captures the context features, this paper proposes a self-attention mechanism to further capture important information. It can better assign weight to important information, and more accurately understand sequence semantics. The calculation of the self-attention mechanism is shown in figure 4.
\begin{figure}[htbp]
\centering\includegraphics[width=0.3\textwidth]{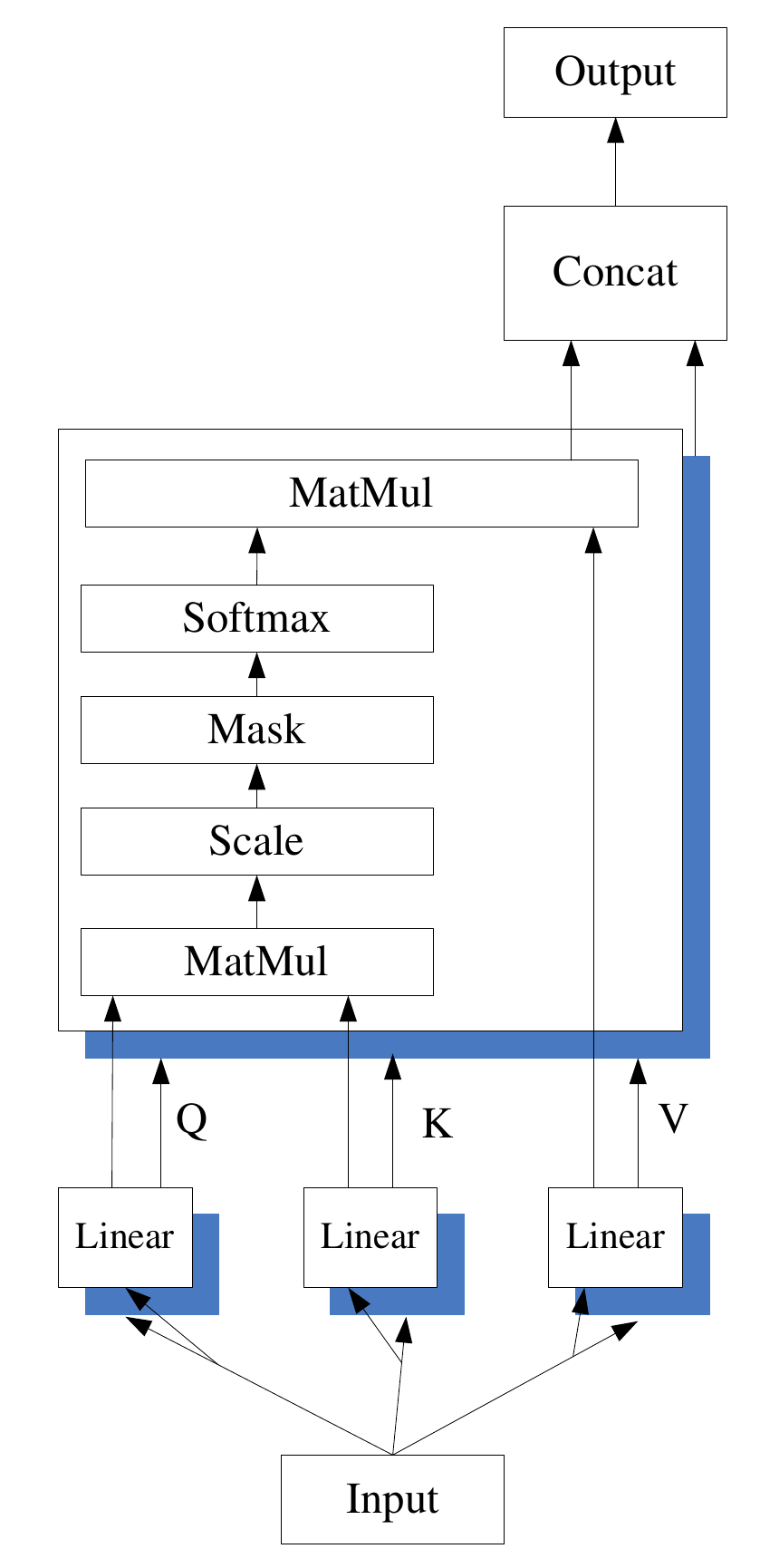}
\caption{Self-attention mechanism calculation method}
\label{fig4}
\end{figure}

Self-attention mechanism is regarded as a mapping of query to a series of key-value pairs, On this basis, considering that an self-attention mechanism cannot capture important features from multiple angles and multiple levels, it is necessary to use the multi-head attention mechanism. The multi-head attention mechanism maps the input into multiple vector spaces and computes the context representation of the character in the vector space, repeating the process several times, and finally stitching the results together. Get a comprehensive character context feature. The calculation formula of multi-head attention mechanism is as follows:
\begin{eqnarray}
\begin{aligned}
m(\mathop {{h_t}}) = concat(scor{e_1}(\mathop {{h_t}}),scor{e_2}(\mathop {{h_t}}),\\
...,scor{e_h}(\mathop {{h_t}})){W^O}
\end{aligned}
\end{eqnarray}

Among them, $\mathop {{h_t}}$ is the hidden layer output of BiGRU, ${score_i}$ is the output of the ${i^{th}}$ self-attention mechanism, ${h}$ is repeat number, ${score_i}$ is calculated as follows:
\begin{eqnarray}
scor{e_i}(\mathop {{h_t}}) = attention(\mathop {{h_t}}  {W_i}^Q,\mathop {{h_t}}  {W_i}^K,\mathop {{h_t}}  {W_i}^V)
\end{eqnarray}

Among them, ${W_i}^Q$, ${W_i}^K$, ${W_i}^V$ and ${W}^O$ are parameter matrices, which are used to map input $\mathop {{h_t}}$ into different vector spaces. The parameter matrix sizes are ${W_i}^Q \in {R^{d \times {d_Q}}}$, ${W_i}^K \in {R^{d \times {d_Q}}}$, ${W_i}^V \in {R^{d \times {d_V}}}$, ${W^O} \in {R^{h{d_V} \times d}}$ respectively. Where $d$ is the output vector dimension of BiGRU network hidden layer, ${d_Q}$ and ${d_V}$ are vector space dimensions. Function attention is self-attention mechanism operation of the dilated dot product, and the formula is defined as:
\begin{eqnarray}
attention(Q,K,V) = {softmax}(\frac{{Q{K^T}}}{{\sqrt d }})V
\end{eqnarray}

Where $\sqrt{d}$ is play the role of scaling adjustment, so that the inner product is not too large. The hidden layer output of the BiGRU network at time $t$ is defined as ${H_t} = m(h_{t})$. The output sequence of the BiGRU network is $H = (H_{1},H_{2},...,H_{n})$, which is also the output of this layer.
\subsection{CapsNet Entity Recognition Layer}
In the field of entity recognition, most recognition methods mapping output features of BiGRU hidden layer to entity label probability vectors through weight matrix, and use scalar to represent entity label prediction probability. In this paper, CapsNet is introduced to recognize entities, in which capsule represents the entity label, the modulus length of capsule vector represents the entity label prediction probability, and the direction of capsule vector represents the attribute of the entity. Since capsule uses vector representation instead of scalar representation, it has stronger ability to express entity information.

We illustrate our proposed CapsNet in figure 5. For feature output from BiGRU and self-attention mechanism layer, the feature dimension is fixed by the matrix $W$ and the primary capsule layer is constructed. The number of primary capsule is set 32. The CapsNet is divided into two layers: primary capsule layer and digit capsule layer. There are connection weights between the two layers. The connection weight will change during the network training, which is called dynamic routing. The pseudo-code of the dynamic routing algorithm is shown in Algorithm 1.
\begin{figure}
\centering
\includegraphics[width=0.45\textwidth]{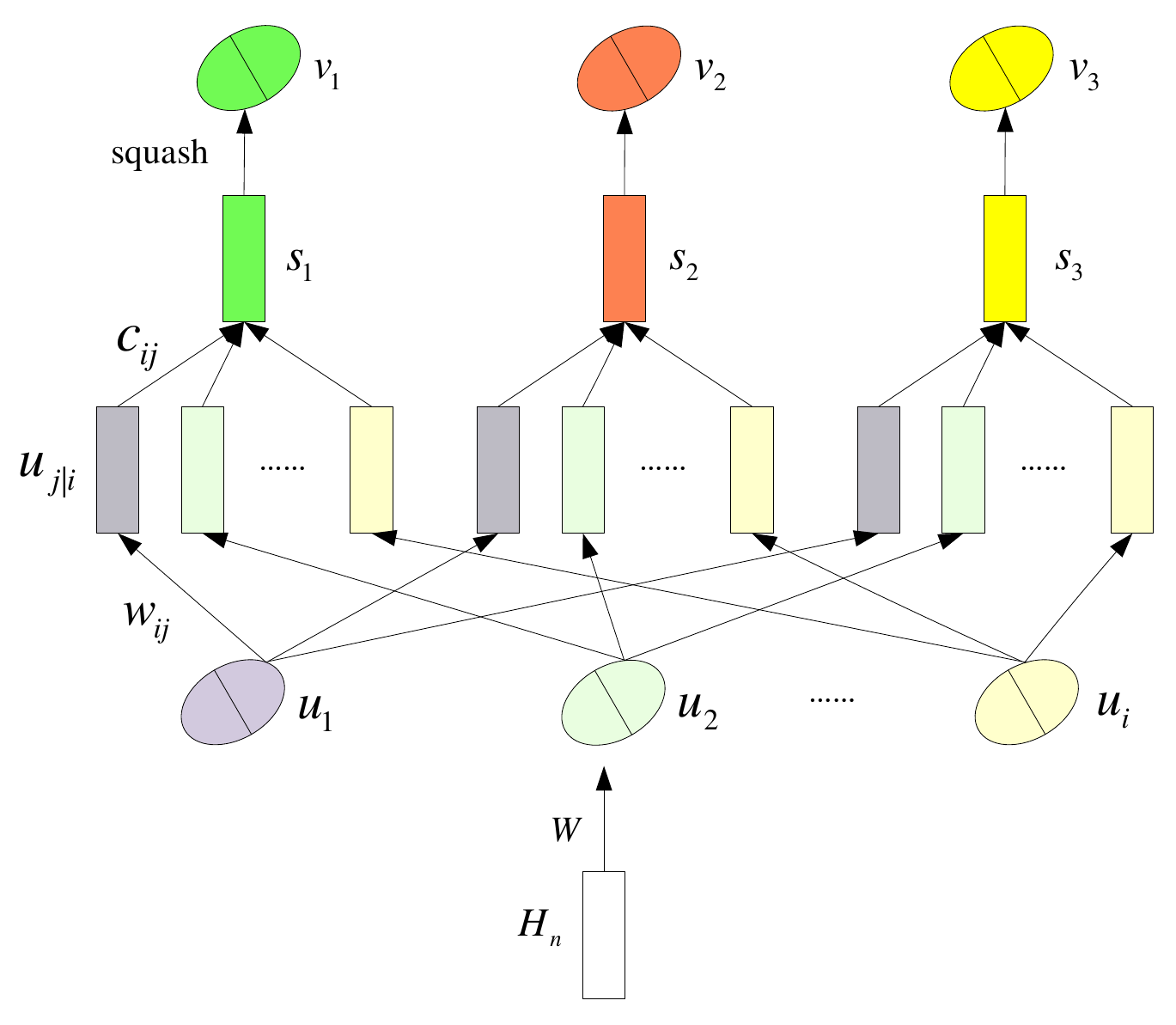}
\caption{An Example of CapsNet entity category classification}
\label{fig5}
\end{figure}

 \begin{algorithm}[htb]  
  \caption{Dynamic routing algorithm.}  
  \label{alg:Framwork}  
  \begin{algorithmic}[1]  
    \Require  
      ${u_{j|i}}$;  
      ${r}$;
    \Ensure  
      $v_j$;  
    \State for all primary capsule $i$ and digit capsule $j$ : ${b_{ij}} = 0$;  
    \For {$r$ $iterations$};  
      \For {all primary capsule $i$ and digit capsule $j$};
        \State ${c_{ij}} \leftarrow softmax ({b_{ij}})$;
      \EndFor
      \For {all digit capsule $j$};
        \State ${s_j} \leftarrow \sum\limits_i {{c_{ij}}{u_{j|i}}} $;
      \EndFor
      \For {all digit capsule $j$};
        \State ${v_j} \leftarrow squash({s_j})$;
      \EndFor
      \For {all primary capsule $i$ and digit capsule $j$};
        \State ${b_{ij}} \leftarrow {b_{ij}} + {u_{j|i}} \cdot {v_j}$;
      \EndFor
    \EndFor
    \State \Return $v_j$;
  \end{algorithmic}  
\end{algorithm}

As shown in Algorithm 1, we know that vector is $u_{j|i}$. $u_{j|i}$ represents the predictive vector of the output of the ${i^{th}}$ primary capsule to the ${j^{th}}$ digit capsule. It is obtained by the product of the primary capsule output vector ${u_i}$ and the weight matrix ${W_{ij}}$, as shown in formula 12. ${b_{ij}}$ is the logarithmic prior probability of the ${i^{th}}$ capsule and the ${j^{th}}$ capsule, initialized to 0. In the iterative process, firstly, ${b_{ij}}$ is normalized by softmax to obtain ${c_{ij}}$, which represents the probability that the ${i^{th}}$ capsule connects to the ${j^{th}}$ capsule. Secondly, 
sum the products of all predictive vectors and their corresponding connection probabilities to obtain digital capsule input ${s_j}$, as shown in formula 13. Apply the squashing function to ${s_j}$ to obtain the digit capsule output ${v_j}$, the squashing function is shown in formula 14. Finally update the corresponding weight ${b_{ij}}$ according to formula 15, and repeat the process until convergence.
\begin{eqnarray}
{u_{j|i}} = {W_{ij}} \cdot {u_i}
\end{eqnarray}
\begin{eqnarray}
{s_j} = \sum\limits_i {{c_{ij}}{u_{j|i}}}
\end{eqnarray}
\begin{eqnarray}
{v_j} = \frac{{{{\left\| {{s_j}} \right\|}^2}}}{{1 + {{\left\| {{s_j}} \right\|}^2}}}\frac{{{s_j}}}{{\left\| {{s_j}} \right\|}}
\end{eqnarray}
\begin{eqnarray}
{b_{ij}} = {b_{ij}} + {u_{j|i}} \cdot {v_j}
\end{eqnarray}

After the entity is recognized by the CapsNet, the sentence sequence is labelled with entity labels. Because the labeling of each word in the sentence has a strong dependency, the labeling of one word affects the labeling information of the next word, so this paper introduces a markov transition matrix to represent the effects between entity labels. For constraints, the markov transition matrix is denoted as $A$, and the transition probability ${A_{ij}}$ is the probability of transition from the ${i^{th}}$ label to the ${j^{th}}$ label. The entity recognition result depends on two results: the recognition result of the CapsNet and the transfer result of the label. Given the character feature sequence $x=(x_{1},x_{2},...,x_{n})$, its corresponding sequence of entity labels is $y=(y_{1},y_{2},...,y_{n})$. Sequence evaluation score is shown in formula 16.
\begin{eqnarray}
S(x,y) = \sum\limits_{i = 1}^n {({A_{{y_{(i - 1)}}{y_i}}} + score(i,{y_i})} ) 
\end{eqnarray}

Among them, ${A_{{y_{(i - 1)}}{y_i}}}$ represents the probability that the entity label of the previous character is transferred to the entity label of the current character. $score(i,{y_i})$ represents the capsule modulus length corresponding to the current character labeled as the ${y_i^{th}}$ entity label. In model training, set the number of possible label sequences to $m$. The probability of labeling sequence ${y}$ is shown in formula 17.
\begin{eqnarray}
p(y|x) = \frac{e^{S(x,y)}}{\sum\limits_{i = 1}^m e^{S(x,y_{i}^{'})}}
\end{eqnarray}

The maximum likelihood method is used to solve the maximum posterior probability of ${p(y|x)}$, as shown in formula 18. When the value of formula 18 reaches its maximum, an optimal labeled sequence is found, which is the optimal entity recognition result for the given sentence.
\begin{eqnarray}
log(p(y|x))= S(x,y)-log(\sum\limits_{i = 1}^m e^{S(x,y_{i}^{'})})
\end{eqnarray}

In summary, B-SABCN model for text classification implementation process is shown as Algorithm 2.
 \begin{algorithm}[htb]  
  \caption{Pseudo-code for B-SABCN.}  
  \label{alg:Framwork2}  
  \begin{algorithmic}[1]  
    \Require  
      $s=(c_1,c_2,...,c_n)$;  
    \Ensure  
      $y=(y_1,y_2,...,y_n)$;  
    \State The character embedding sequence representation generated by the BERT pre-training method is shown as $x=(x_1,x_2,...,x_n)$;  
    \State Use formula (6) to calculate the output of the forward hidden layer of the GRU unit $\mathop {{h_t}}\limits^ \to$, and then get the forward hidden layer output of the BiGRU as $\mathop {{h}}\limits^ \to=(\mathop {{h_1}}\limits^ \to,\mathop {{h_2}}\limits^ \to,...,\mathop {{h_n}}\limits^ \to)$;
    \State Use formula (7) to calculate the output of the backward hidden layer of the GRU unit $\mathop {{h_t}}\limits^ \leftarrow$, and then get the backward hidden layer output of the BiGRU as $\mathop {{h}}\limits^ \leftarrow=(\mathop {{h_1}}\limits^ \leftarrow,\mathop {{h_2}}\limits^ \leftarrow,...,\mathop {{h_n}}\limits^ \leftarrow)$;
    \State Use formula (8) to splice the forward and backward hidden layer output of BiGRU to get the hidden layer output of BiGRU as $h=(h_1,h_2,...,h_n)$.
    \State Use formula (9)-(11) to further capture the context related information of the sentence sequence from multiple angles, and assign different weights to different characters. The output is shown as $H=(H_1,H_2,...,H_n)$; 
    \State Feed $H$ into the CapNets to get the entity label probability of characters, and introduce a markov probability matrix to represent the entity label transition probability. Use the formula (16) to get the sequence entity label score. Maximize the formulas (18) to get an optimal entity label sequence; 
    \State \Return $y=(y_1,y_2,...,y_n)$;
  \end{algorithmic}  
\end{algorithm}
\section{Experiments}
\subsection{Datasets}
In order to evaluate the model proposed in this paper based on Chinese NER data, we conducted experiments on MSRA entity recognition data\cite{levow2006third} and diabetes entity recognition data provided by ruijin hospital\cite{Alibaba}. MSRA data set is a simplified Chinese entity identification data set published by Microsoft, which includes three entity types: PER(person), ORG(organization) and LOC(location). This data set does not have a validation set, so we extract 10$\% $ of the training set data as the validation set.

The diabetes entity recognition data set of ruijin hospital is from authoritative journals in the field of diabetes in Chinese, and the annotators all have medical backgrounds. The data include basic research, clinical research, drug use, clinical cases, diagnosis and treatment methods, etc. This paper preprocesses the original data, including 15 entities: Disease, Reason, Symptom, Test, Test$\_$Value, Drug, Frequency, Amount, Method, Treatment, Operation, SideEff, Anatomy, level, Duration. Since there is no test set for diabetes data, we extracted 10$\% $ of the training set data as the test set.
\subsection{Model setting}
For evaluation, Precision(P), Recall(R) and F1 score were used as the indexes of our experiment. For model parameter configuration, the embedding size of characters is set to 768, the hidden state dimension of GRU unit is set to 100, the initial learning rate is set to 0.001, and the dropout rate is set to 0.5. We use the BIOES tagging scheme to process entity labels\cite{ratinov2009design}. Each character in the text is processed as B(begin), I(inside), O(outside), E(end), S(single). Table \uppercase\expandafter{\romannumeral1} shows model parameter configuration.
\begin{table}
\centering
\caption{Experimental model parameters}
\label{tab:1}       
\begin{tabular}{cc}
\hline\noalign{\smallskip}
Parameter names & Values   \\
\noalign{\smallskip}\hline\noalign{\smallskip}
Character vector dimension&
768\\
GRU unit dimension&
100\\
Capsule vector dimension&
16\\
Dropout rate&
0.5\\
Learning rate&
0.001\\
Optimizer&
Adam\\
Batch size&
20\\
Tag schema&
BIOES\\
\noalign{\smallskip}\hline
\end{tabular}
\end{table}
\subsection{Experimental Results}
In this section, we present the experimental results of the proposed model on the MSRA dataset and the diabetes dataset, and compare it with the latest methods available.
\begin{enumerate}
\item \emph {Evaluation on MSRA datasets.}
Table \uppercase\expandafter{\romannumeral2} shows the experimental results of B-SABCN model in this paper and the latest model before Chinese NER on MSRA dataset. Dong et al.\cite{dong2016character} uses character level and radical level word embedding representations as input to the LSTM-CRF model. Yang et al.\cite{yang2018five} proposed CNN-BiRNN-CRF entity recognition model based on five-stroke. Cao et al.\cite{cao2018adversarial} proposed joint training Chinese entity recognition model based on adversarial transfer learning and attention mechanism. Zhang et al.\cite{zhang2018chinese} constructed a lattice LSTM model and integrated the potential word information of characters into the neural network by using external dictionary data. Zhang et al.\cite{zhang2019chinese} studied a dynamic meta embedding method. Zhu et al.\cite{zhu2019can}  proposed the convolution attention network model. Jin et al.\cite{jin2019lstm} proposed the convolutional recursive neural network model and applied it to Chinese NER task.
\begin{table}
\centering
\caption{Experimental results on MSRA dataset}
\label{tab:2}       
\begin{tabular}{cccc}
\hline\noalign{\smallskip}
Model&
P&
R&
F1 score   \\
\noalign{\smallskip}\hline\noalign{\smallskip}
Dong et al.\cite{dong2016character}&
91.28\%&
90.62\%&
90.95\%\\
Yang et al.\cite{yang2018five}&
92.04\%&
91.31\%&
91.67\%\\
Cao et al.\cite{cao2018adversarial}&
91.30\%&
89.58\%&
90.64\%\\
Zhang et al.\cite{zhang2018chinese}&
93.57\%&
92.79\%&
93.18\%\\
Zhang et al.\cite{zhang2019chinese}&
90.59\%&
91.15\%&
90.87\%\\
Zhu et al.\cite{zhu2019can}&
93.53\%&
92.42\%&
92.97\%\\
Jin et al.\cite{jin2019lstm}&
93.71\%&
92.46\%&
93.08\%\\
B-SABCN&
95.12\%&
94.82\%&
94.97\%\\
\noalign{\smallskip}\hline
\end{tabular}
\end{table}

\setlength{\parindent}{1em} As can be seen from the Table \uppercase\expandafter{\romannumeral2}, B-SABCN model proposed in this paper does not need to provide external dictionary information, and its overall performance F1 value reaches 94.97$\% $, which is higher than other existing models. The experimental results also prove that the introduction of BERT pre-trained language models can better represent the semantic information of characters. The self-attention mechanism proposed in this paper gives different focus on the forward and backward hidden layers of BiGRU respectively, and fully capture the context related information in the sentence sequence. The introduction of CapsNet improves B-SABCN model's entity information representation capabilities, so that they can improve text entity recognition performance.
\item \emph {Evaluation on Diabetes datasets.}
In the diabetes dataset, since this dataset is a competition dataset, the B-SABCN model is compared with the mainstream models of BiLSTM-CRF, IDCNN-CRF and CNN-BiLSTM-CRF. The model of BiLSTM-CRF mainly uses word2vec to train character vectors, then combined with BiLSTM network for training, finally the trained feature matrix input to CRF for obtaining the optimal entity label sequence. IDCNN-CRF is similar to BiLSTM-CRF, except that BiLSTM is replaced with IDCNN, the others remain unchanged. CNN-BiLSTM-CRF is used by the competition champion, which uses word2vec to train the character vector, and uses CNN to obtain the local features of the sequence. The local features and the character vector are combined as the input of BiLSTM network, and the trained feature matrix input to CRF for obtaining the optimal entity label sequence. The comparison results are shown in Table \uppercase\expandafter{\romannumeral3}.
\begin{table}
\centering
\caption{Experimental results on Diabetes dataset}
\label{tab:3}       
\begin{tabular}{cccc}
\hline\noalign{\smallskip}
Model&
P&
R&
F1 score   \\
\noalign{\smallskip}\hline\noalign{\smallskip}
BiLSTM-CRF&
72.47\%&
71.61\%&
72.04\%\\
IDCNN-CRF&
73.07\%&
70.15\%&
71.56\%\\
CNN-BiLSTM-CRF&
-&
-&
76.3\%\\
B-SABCN&
78.29\%&
78.03\%&
78.16\%\\
\noalign{\smallskip}\hline
\end{tabular}
\end{table}

As shown in Table \uppercase\expandafter{\romannumeral3}, our B-SABCN method has obvious improvements. Compared with the IDCNN-CRF model, BiGRU can learn more context features. Compared with the BiLSTM-CRF model, the B-SABCN model increases F1 score by 6$\% $. Compared with the CNN-BiLSTM-CRF model, although CNN can capture local features of the character sequence, but ignore the important information of capture sequence, the B-SABCN model introduces self-attention mechanism and CapsNet to enhance the ability to capture important information and achieve better entity recognition results.
\item \emph {Evaluation on the components of B-SABCN model.}
In this set of experiments, the B-SABCN model is divided into the following situations:

(1)The word2vec model is used to train the character vector as the input of the B-SABCN model, called W-SABCN. It tests the contribution of the BERT pre-trained model to the B-SABCN model.

(2)The B-SABCN model lacks a self-attention mechanism, called B-BCN. It tests the contribution of the self-attention mechanism to the B-SABCN model. 

(3)The B-SABCN model uses scalars to recognize entities, called B-SABN, and tests the contribution of the CapsNet to the model. 

Compare the models in these three cases with the B-SABCN model. In addition, word2vec-BiGRU-CRF is introduced as the baseline method, and BERT-BiGRU-CRF is used as BERT-Baseline method. The comparison results of the two data sets are shown in Table \uppercase\expandafter{\romannumeral4}.
\begin{table}
\centering
\caption{Comparison of each part contribution in B-SABCN model on MSRA dataset and diabetes dataset}
\label{tab:4}
\setlength{\tabcolsep}{3pt}
\begin{tabular}{ccccccc}
\hline\noalign{\smallskip}
\multirow{2}*{Model} &
\multicolumn{3}{c}{MSRA dataset} &
\multicolumn{3}{c}{Diabetes dataset}\\
\cline{2-7}\noalign{\smallskip}
& F & R & F1 score & F & R & F1 score\\
\hline\noalign{\smallskip}
Baseline&
90.14\%&
89.40\%&
89.77\%&
72.26\%&
71.49\%&
71.87\%\\
W-SABCN&
91.37\%&
90.48\%&
90.92\%&
74.51\%&
73.17\%&
73.84\%\\
BERT-Baseline&
94.18\%&
94.06\%&
94.12\%&
76.81\%&
77.05\%&
76.93\%\\
B-BCN&
94.50\%&
94.30\%&
94.40\%&
77.37\%&
77.25\%&
77.31\%\\
B-SABN&
94.74\%&
94.57\%&
94.66\%&
77.84\%&
77.60\%&
77.72\%\\
B-SABCN&
95.12\%&
94.82\%&
94.97\%&
78.29\%&
78.03\%&
78.16\%\\
\hline
\end{tabular}
\end{table}

From the results in Table \uppercase\expandafter{\romannumeral4}, it can be seen that in case (1), the F1 score of the W-SABCN model using traditional character embedding dropped the most, reaching 4$\% $, which indicates that the BERT character vector embedding in the B-SABCN model can dynamically generate character vectors based on context related information. Accurately understand the semantics and overcome polysemy problem of word2vec.

In case (2), the lack of self-attention mechanism will also reduce the value of model F1, indicating that the self-attention mechanism can fully capture useful features in the text itself, calculate the correlation between any two character features in the sentence, and better understand the sentence structure. Therefore, the entity that could not be recognized originally can be recognized correctly.

In case (3), the declining degree of the F1 value of the model that uses scalars for entity classification is smaller than the previous two, which proves that the capsule network uses vectors to represent entity categories, and the vector modulus length represents the probability of entity categories to obtain better recognition results.

Finally, the effect of the B-SABCN model is the best when BERT, self-attention mechanism and capsule network are included, which verifies the effectiveness of the proposed method in improving the performance of entity recognition.
\end{enumerate}
\section{Conclusion}
In this paper, we propose a Chinese NER task model. The model generates character vectors through BERT pre-training language model, and splices the position vectors of characters in words with the character vectors as a new character vector feature representation. In our model, we use self-attention mechanism to capture the deep information of the hidden layer of BiGRU network from multiple vector spaces. Finally, the CapsNet is introduced to classify entity categories. Experiments on two data sets show that the proposed model can achieve better performance without providing external resource information. 

In the future, we can consider using transfer learning method to improve the entity recognition performance of small scale domain data sets by using large scale public data sets.
\section*{Acknowledgment}

The preferred spelling of the word ``acknowledgment'' in America is without 
an ``e'' after the ``g''. Avoid the stilted expression ``one of us (R. B. 
G.) thanks $\ldots$''. Instead, try ``R. B. G. thanks$\ldots$''. Put sponsor 
acknowledgments in the unnumbered footnote on the first page.

\bibliographystyle{./bibliography/IEEEtran}
\bibliography{./bibliography/ref}
\end{document}